\documentclass[sigconf]{acmart}

\AtBeginDocument{%
  \providecommand\BibTeX{{%
    \normalfont B\kern-0.5em{\scshape i\kern-0.25em b}\kern-0.8em\TeX}}}

\setcopyright{acmcopyright}
\copyrightyear{2019}
\acmYear{2019}

\acmConference[WISDOM@KDD'19]{Proceedings of 8$^{th}$ KDD Workshop on Issues of Sentiment Discovery and
Opinion Mining (WISDOM)}{August 04--08, 2019}{Anchorage, Alaska, USA}
\acmBooktitle{Proceedings of 8$^{th}$ KDD Workshop on Issues of Sentiment Discovery and Opinion Mining (WISDOM@KDD'19), August 04--08, 2019, Anchorage, Alaska, USA}

\usepackage{hyperref}
\usepackage{graphicx}
\usepackage{subcaption}
\usepackage{comment}
\usepackage{url}
\usepackage{multirow}
\usepackage{adjustbox}

\begin{document}

\title{Yoga-Veganism: Correlation Mining of Twitter Health Data}

\author{Tunazzina Islam}
\orcid{0000-0001-6714-973X}
\affiliation{%
  \institution{Department of Computer Science \\ Purdue University, West Lafayette}
  \state{IN-47907, USA}
}
\email{islam32@purdue.edu}

\begin{abstract}
Nowadays social media is a huge platform of data. People usually share their interest, thoughts via discussions, tweets, status. It is not possible to go through all the data manually. We need to mine the data to explore hidden patterns or unknown correlations, find out the dominant topic in data and understand people's interest through the discussions. In this work, we explore Twitter data related to health. We extract the popular topics under different categories (e.g. diet, exercise) discussed in Twitter via topic modeling, observe model behavior on new tweets, discover interesting correlation (i.e. Yoga-Veganism). We evaluate accuracy by comparing with ground truth using manual annotation both for train and test data.
\end{abstract}


\maketitle

\section{Introduction}
The main motivation of this work has been started with a question "What do people do to maintain their health?"-- some people do balanced diet, some do exercise. Among diet plans some people maintain vegetarian diet/vegan diet, among exercises some people do swimming, cycling or yoga. There are people who do both. If we want to know the answers of the following questions-- "How many people follow diet?", "How many people do yoga?", "Does yogi follow vegetarian/vegan diet?", may be we could ask our acquainted person but this will provide very few intuition about the data. Nowadays people usually share their interests, thoughts via discussions, tweets, status in social media (i.e. Facebook, Twitter, Instagram etc.). It's huge amount of data and it's not possible to go through all the data manually. We need to mine the data to get overall statistics and then we will also be able to find some interesting correlation of data.

Several works have been done on prediction of social media content \cite{de2013predicting}, \cite{cobb2012health}, \cite{eichstaedt2018facebook}, \cite{reece2017forecasting}, \cite{son2017recognizing}. Prieto et al. proposed a method to extract a set of tweets to estimate and track the incidence of health conditions in society \cite{prieto2014twitter}. Discovering public health topics and themes in tweets had been examined by Prier et al. \cite{prier2011identifying}. Yoon et al. described a practical approach of content mining to analyze tweet contents and illustrate an application of the approach to the topic of physical activity \cite{yoon2013practical}.

Twitter data constitutes a rich source that can be used for capturing information about any topic imaginable. In this work, we use text mining to mine the Twitter health-related data. Text mining is the application of natural language processing techniques to derive relevant information \cite{allahyari2017brief}. Millions of tweets are generated each day on multifarious issues \cite{pandarachalil2015twitter}. Twitter mining in large scale has been getting a lot of attention last few years. Lin and Ryaboy discussed the evolution of Twitter infrastructure and the development of capabilities for data mining on "big data" \cite{lin2013scaling}. Pandarachalil et al. provided a scalable and distributed solution using Parallel python framework for Twitter sentiment analysis \cite{pandarachalil2015twitter}. Large-scale Twitter Mining for drug-related adverse events was developed by Bian et al. \cite{bian2012towards}.

In this paper, we use parallel and distributed technology Apache Kafka \cite{kreps2011kafka} to handle the large streaming twitter data. The data processing is conducted in parallel with data extraction by integration of Apache Kafka and Spark Streaming. Then we use Topic Modeling to infer semantic structure of the unstructured data (i.e Tweets). Topic Modeling is a text mining technique which automatically discovers the hidden themes from given documents. It is an unsupervised text analytic algorithm that is used for finding the group of words from the given document. We build the model using three different algorithms Latent Semantic Analysis (LSA) \cite{deerwester1990indexing}, Non-negative Matrix Factorization (NMF) \cite{lee2001algorithms}, and Latent Dirichlet Allocation (LDA) \cite{blei2003latent} and infer the topic of tweets. To observe the model behavior, we test the model to infer new tweets. The implication of our work is to annotate unlabeled data using the model and find interesting correlation.

\section{Data Collection}
Tweet messages are retrieved from the Twitter source by utilizing the Twitter API and stored in Kafka topics. The Producer API is used to connect the source (i.e. Twitter) to any Kafka topic as a stream of records for a specific category. We fetch data from a source (Twitter), push it to a message queue, and consume it for further analysis. Fig. \ref{fig:data_collection} shows the overview of Twitter data collection using Kafka.

\subsection{Apache Kafka} 
\begin{figure}[htbp]
  \centering  
  \includegraphics[width= 0.5\textwidth]{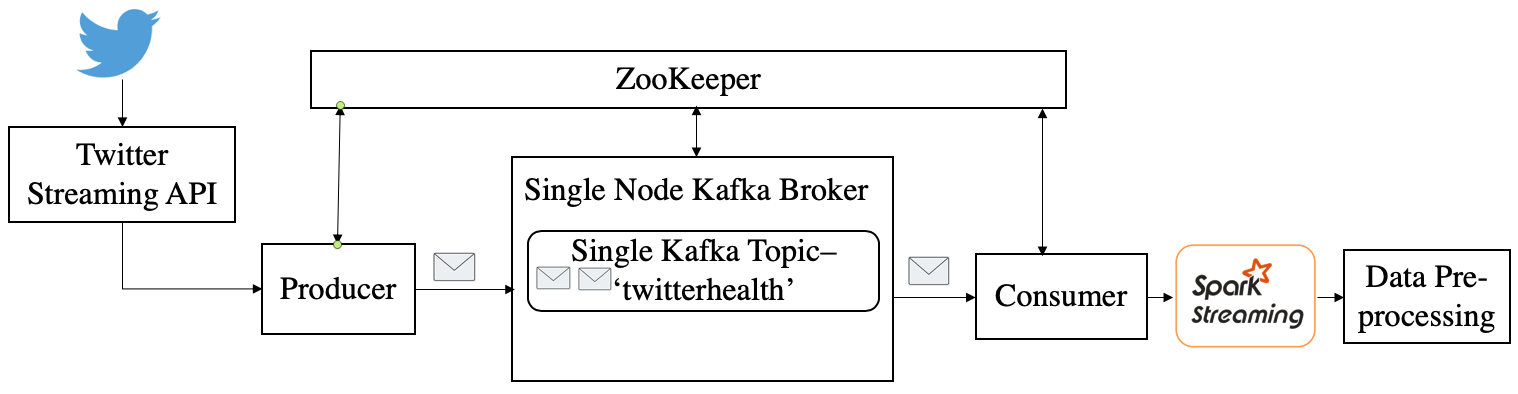}
    \caption{Twitter Data Collection.}
    \label{fig:data_collection}
\end{figure}

In order to handle the large streaming twitter data, we use parallel and distributed technology for big data framework. In this case, the output of the twitter crawling is queued in messaging system called Apache Kafka. This is a distributed streaming platform created and open sourced by LinkedIn in 2011 \cite{kreps2011kafka}. We write a Producer Client which fetches latest tweets continuously using Twitter API and push them to single node Kafka Broker. There is a Consumer that reads data from Kafka (Fig. \ref{fig:data_collection}).

\subsection{Apache Zookeeper}
Apache Zookeeper is a distributed, open-source configuration, synchronization service along with naming registry for distributed applications. Kafka uses Zookeeper to store metadata about the Kafka cluster, as well as consumer client details. 

\subsection{Data Extraction using Tweepy} 
The twitter data has been crawled using Tweepy which is a Python library for accessing the Twitter API. We use Twitter streaming API to extract 40k tweets (April 17-19, 2019). For the crawling, we focus on several keywords that are related to health. The keywords are processed in a non-case-sensitive way. We use filter to stream all tweets containing the word `yoga', `healthylife', `healthydiet', `diet',`hiking', `swimming', `cycling', `yogi', `fatburn', `weightloss', `pilates', `zumba', `nutritiousfood', `wellness', `fitness', `workout', `vegetarian', `vegan', `lowcarb', `glutenfree', `calorieburn'. 

The streaming API returns tweets, as well as several other types of messages (e.g. a tweet deletion notice, user update profile notice, etc), all in JSON format. We use Python libraries json for parsing the data, pandas for data manipulation.


\subsection{Data Pre-processing}
Data pre-processing is one of the key components in many text mining
algorithms \cite{allahyari2017brief}. Data cleaning is crucial for generating a useful topic model. We have some prerequisites i.e. we download the stopwords from NLTK (Natural Language Toolkit) and spacy's en model for text pre-processing. 

It is noticeable that the parsed full-text tweets have many emails, `RT', newline and extra spaces that is quite distracting. We use Python Regular Expressions (re module) to get rid of them. Then we tokenize each text into a list of words, remove punctuation and unnecessary characters. We use Python Gensim package for further processing. Gensim's simple\_preprocess() is used for tokenization and removing punctuation. We use Gensim's Phrases model to build bigrams. Certain parts of English speech, like conjunctions ("for", "or") or the word "the" are meaningless to a topic model. These terms are called stopwords and we remove them from the token list. We use spacy model for lemmatization to keep only noun, adjective, verb, adverb. Stemming words is another common NLP technique to reduce topically similar words to their root. For example, "connect", "connecting", "connected", "connection", "connections" all have similar meanings; stemming reduces those terms to "connect". The Porter stemming algorithm \cite{porter1980algorithm} is the most widely used method.

\begin{figure}[htbp]
  \centering  
  \includegraphics[width= 0.5\textwidth]{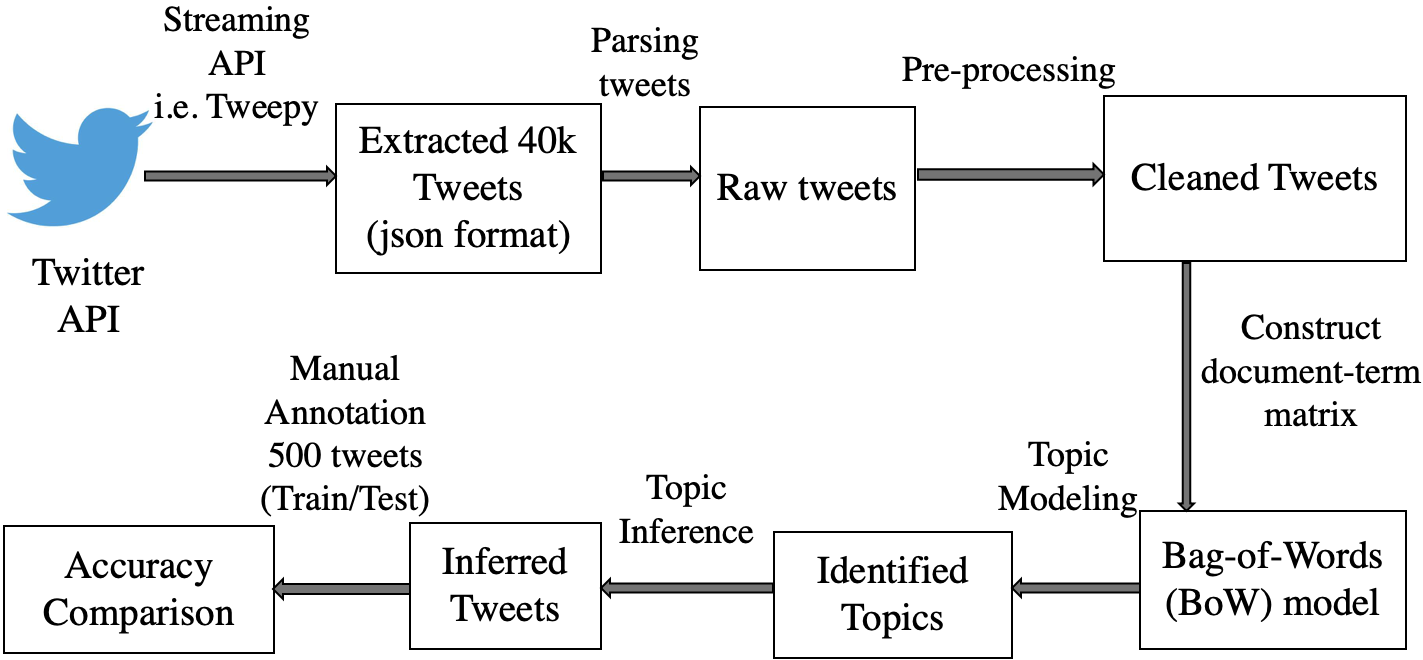}
    \caption{Methodology of correlation mining of Twitter health data.}
    \label{fig:method}
\end{figure}

\section{Methodology}
We use Twitter health-related data for this analysis. In subsections \hyperref[subsec:3.1]{3.1}, \hyperref[subsec:3.2]{3.2}, \hyperref[subsec:3.3]{3.3}, and \hyperref[subsec:3.4]{3.4} elaborately present how we can infer the meaning of unstructured data. Subsection \hyperref[subsec:3.5]{3.5} shows how we do manual annotation for ground truth comparison. Fig. \ref{fig:method} shows the overall pipeline of correlation mining.

\begin{figure}[htbp]
  \centering  
  \includegraphics[width= 0.5\textwidth]{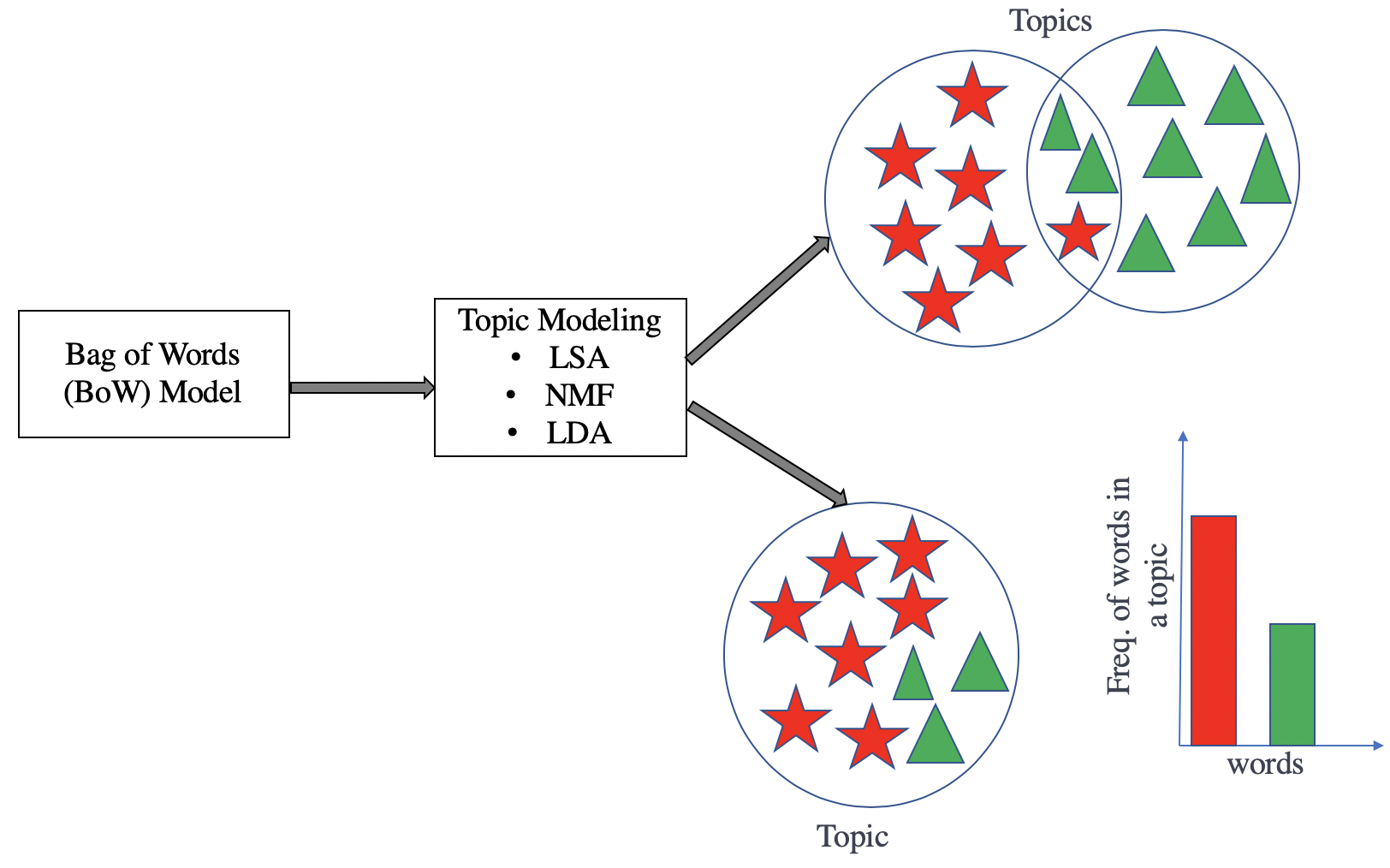}
    \caption{Topic Modeling using LSA, NMF, and LDA. After topic modeling we identify topic/topics (circles). Red pentagrams and green triangles represent group of co-occurring related words of corresponding topic.}
    \label{fig:topic_modeling}
\end{figure}

\begin{figure*}
\centering
    \begin{subfigure}[b]{0.4\textwidth}            
            \includegraphics[width=\textwidth]{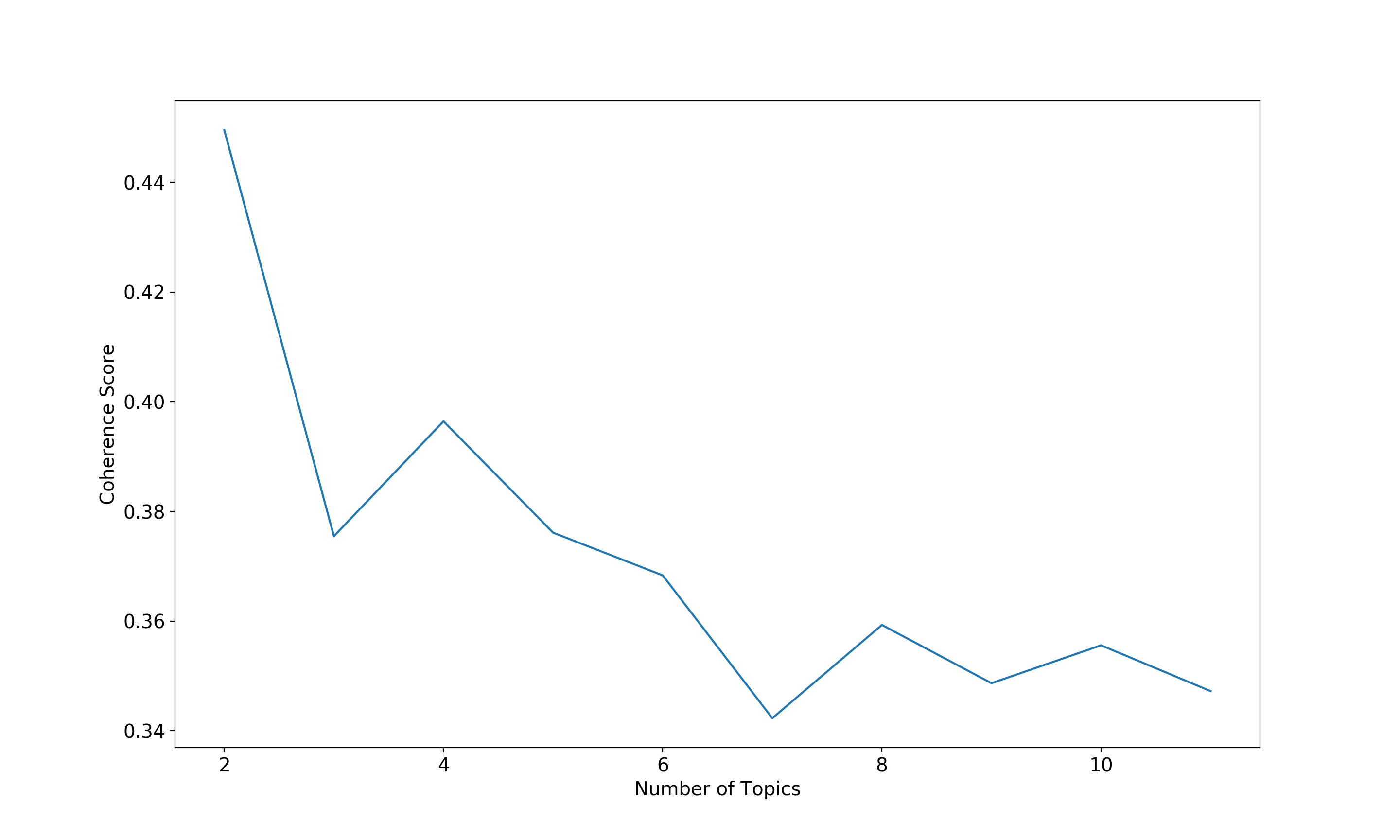}
            \caption{Number of Topics = 2 in LSA.}
            \label{fig:LSA}
    \end{subfigure}
    \begin{subfigure}[b]{0.4\textwidth}
            \centering
            \includegraphics[width=\textwidth]{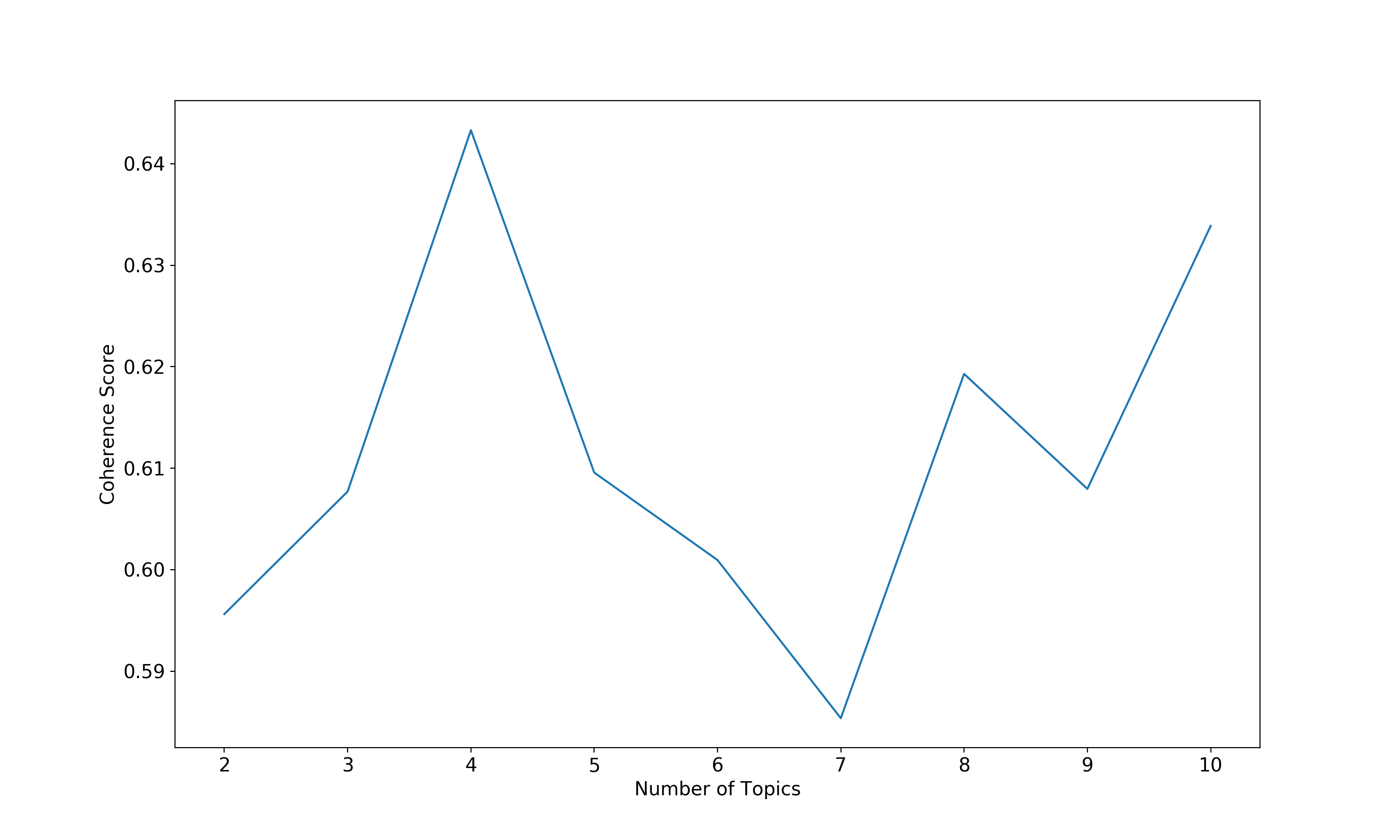}
            \caption{Number of Topics = 4 in NMF.}
            \label{fig:NMF}
    \end{subfigure}
    \begin{subfigure}[b]{0.4\textwidth}
            \centering
            \includegraphics[width=\textwidth]{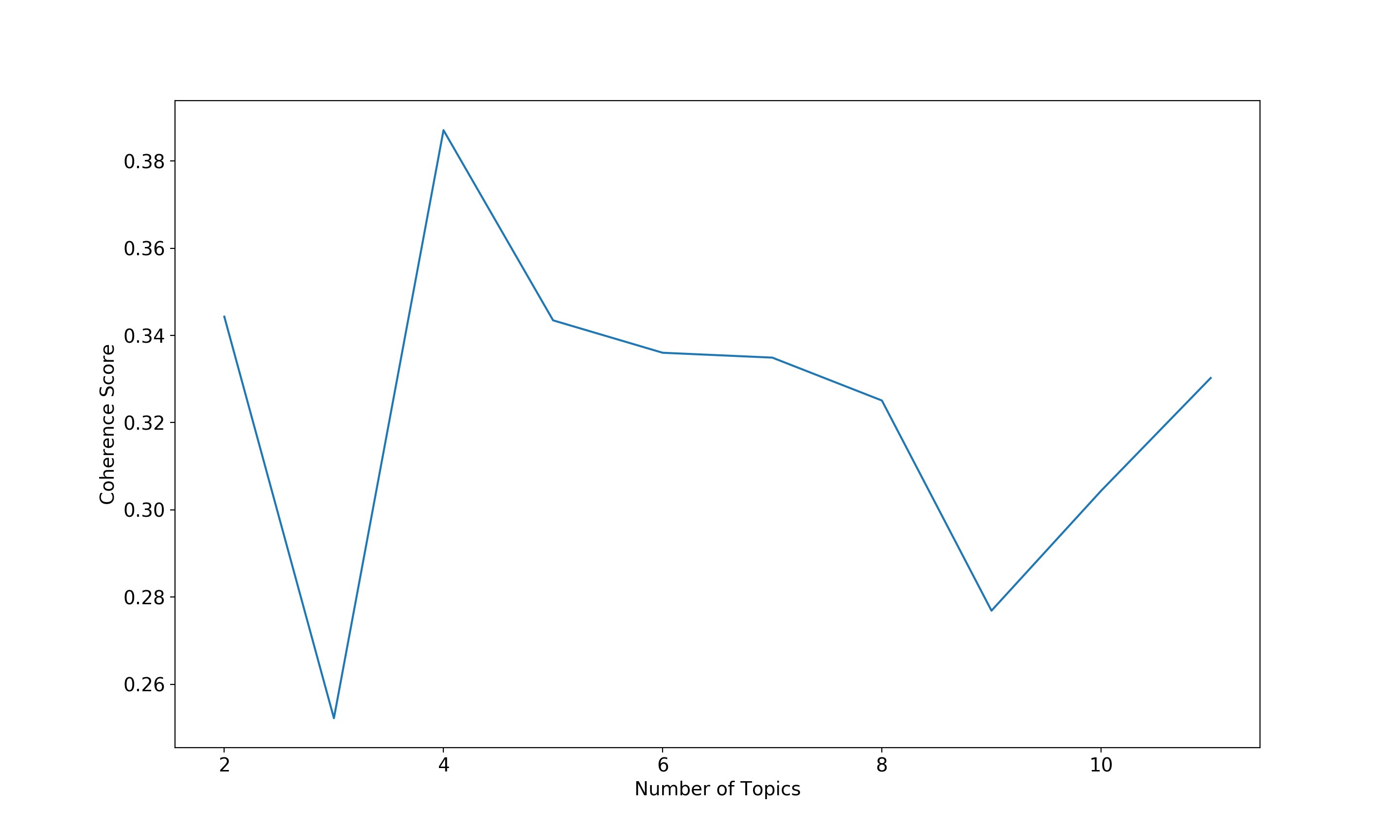}
            \caption{Number of Topics = 4 in LDA.}
            \label{fig:LDA}
    \end{subfigure}
    
    \caption{Optimal Number of Topics vs Coherence Score. Number of Topics (k) are selected based on the highest coherence score. Graphs are rendered in high resolution and can be zoomed in.}\label{fig:k_coherence}
\end{figure*}

\subsection{Construct document-term matrix}
\label{subsec:3.1}
The result of the data cleaning stage is texts, a tokenized, stopped, stemmed and lemmatized list of words from a single tweet. To understand how frequently each term occurs within each tweet, we construct a document-term matrix using Gensim's Dictionary() function. Gensim's doc2bow() function converts dictionary into a bag-of-words. In the bag-of-words model, each tweet is represented by a vector in a m-dimensional coordinate space, where m is number of unique terms across all tweets. This set of terms is called the corpus vocabulary.

\begin{table*}[htbp]
  \centering
  \caption{Topics and top-10 keywords of the corresponding topic}
    \resizebox{\textwidth}{!}{\begin{tabular}{|c|c||c|c|c|c||c|c|c|c|}
    \hline
    \multicolumn{2}{|c||}{\textbf{LSA}} & \multicolumn{4}{|c||}{\textbf{NMF}} & \multicolumn{4}{|c|}{\textbf{LDA}} \\
    \hline
    \multicolumn{1}{|c|}{\textit{Topic 1}} & \multicolumn{1}{|c||}{\textit{Topic 2}} & \multicolumn{1}{|c|}{\textit{Topic 1}}& \multicolumn{1}{|c|}{\textit{Topic 2}} & \multicolumn{1}{|c|}{\textit{Topic 3}} & \multicolumn{1}{|c||}{\textit{Topic 4}} & \multicolumn{1}{|c|}{\textit{Topic 1}} & \multicolumn{1}{|c|}{\textit{Topic 2}} & \multicolumn{1}{|c|}{\textit{Topic 3}} & \multicolumn{1}{|c|}{\textit{Topic 4}} \\
    \hline
    Yoga  & diet  & Yoga  & diet  & vegan & fitness & diet  & vegan  & swimming  & fitness  \\
    \hline
    everi & vegan & job   & beyonce & go & workout & workout & yoga  & swim  & amp \\
    \hline
    Life  & fit   & every\_woman & new   & eat   & go    & new   & job   & day   & wellness \\
    \hline
    Job   & day   & cooks\_goe & bitch & make  & good  & go    & every\_woman   & much  & health  \\
    \hline
    Remember & new   & therapy\_remember & ciara\_prayer & food  & amp   & day   & cooks\_goe  & support  & time  \\
    \hline
    goe   & like  & life\_juggl & day   & day   & day   & beyonce & therapy\_remember &  really  & great  \\
    \hline
    Woman & Beyonce & everyone\_birthday & eat   & amp   & yoga  & get   & life\_juggle  & try   & look \\
    \hline
    Everyone & amp   & boyfriend & go    & shit  & health & today  & everyone\_birthday & always & hiking  \\
    \hline
    cook  & eat   & hot   & fat   & meat  & gym   & bitch & eat   & relationship & make \\
    \hline
    therapy & workout & know  & keto  & vegetarian & today & gym   & boyfriend & pool  & love \\
    \hline
    \end{tabular}}
  \label{tab:topic_keywords}%
\end{table*}%

\subsection{Topic Modeling}
\label{subsec:3.2}
Topic modeling is a text mining technique which provides methods for identifying co-occurring keywords to summarize collections of textual information. This is used to analyze collections of documents, each of which is represented as a mixture of topics, where each topic is a probability distribution over words \cite{alghamdi2015survey}. Applying these models to a document collection involves estimating the topic distributions and the weight each topic receives in each document. A number of algorithms exist for solving this problem. We use three unsupervised machine learning algorithms to explore the topics of the tweets: Latent Semantic Analysis (LSA) \cite{deerwester1990indexing}, Non-negative Matrix Factorization (NMF) \cite{lee2001algorithms}, and Latent Dirichlet Allocation (LDA) \cite{blei2003latent}. Fig. \ref{fig:topic_modeling} shows the general idea of  topic modeling methodology. Each tweet is considered as a document. LSA, NMF, and LDA use Bag of Words (BoW) model, which results in a term-document matrix (occurrence of terms in a document). Rows represent terms (words) and columns represent documents (tweets). After completing topic modeling, we identify the groups of co-occurring words in tweets. These group co-occurring related words makes "topics".

\subsubsection{Latent Semantic Analysis (LSA)}
LSA (Latent Semantic Analysis) \cite{deerwester1990indexing} is also known as LSI (Latent Semantic Index). It learns latent topics by performing a matrix decomposition on the document-term matrix using Singular Value Decomposition (SVD) \cite{golub1971singular}. After corpus creation in \hyperref[subsec:3.1]{Subsection 3.1}, we generate an LSA model using Gensim.

\subsubsection{Non-negative Matrix Factorization (NMF)}
Non-negative Matrix Factorization (NMF) \cite{lee2001algorithms} is a widely used tool for the analysis of high-dimensional data as it automatically extracts sparse and meaningful features from a set of non-negative data vectors. It is a matrix factorization method where we constrain the matrices to be non-negative.

We apply Term Weighting with term frequency-inverse document frequency (TF-IDF) \cite{salton1986introduction} to improve the usefulness of the document-term matrix (created in \hyperref[subsec:3.1]{Subsection 3.1}) by giving more weight to the more "important" terms. In Scikit-learn, we can generate at TF-IDF weighted document-term matrix by using TfidfVectorizer. We import the NMF model class from sklearn.decomposition and fit the topic model to tweets.

\subsubsection{Latent Dirichlet Allocation (LDA)}
Latent Dirichlet Allocation (LDA) \cite{blei2003latent} is widely used for identifying the topics in a set of documents, building on Probabilistic Latent Semantic Analysis (PLSI) \cite{hofmann1999probabilistic}. 
LDA considers each document as a collection of topics in a certain proportion and each topic as a collection of keywords in a certain proportion. We provide LDA the optimal number of topics, it rearranges the topics' distribution within the documents and keywords' distribution within the topics to obtain a good composition of topic-keywords distribution.

We have corpus generated in \hyperref[subsec:3.1]{Subsection 3.1} to train the LDA model. In addition to the corpus and dictionary, we provide the number of topics as well. 

\subsection{Optimal number of Topics}
\label{subsec:3.3}
Topic modeling is an unsupervised learning, so the set of possible topics are unknown. To find out the optimal number of topic, we build many LSA, NMF, LDA models with different values of number of topics (k) and pick the one that gives the highest coherence score. Choosing a `k' that marks the end of a rapid growth of topic coherence usually offers meaningful and interpretable topics.

We use Gensim's coherencemodel to calculate topic coherence for topic models (LSA and LDA). For NMF, we use a topic coherence measure called TC-W2V. This measure relies on the use of a word embedding model constructed from the corpus. So in this step, we use the Gensim implementation of Word2Vec \cite{mikolov2013efficient} to build a Word2Vec model based on the collection of tweets.

We achieve the highest coherence score = 0.4495 when the number of topics is 2 for LSA, for NMF the highest coherence value is 0.6433 for K = 4, and for LDA we also get number of topics is 4 with the highest coherence score which is 0.3871 (see Fig. \ref{fig:k_coherence}). 

For our dataset, we picked k = 2, 4, and 4 with the highest coherence value for LSA, NMF, and LDA correspondingly (Fig. \ref{fig:k_coherence}). Table \ref{tab:topic_keywords} shows the topics and top-10 keywords of the corresponding topic. We get more informative and understandable topics using LDA model than LSA. LSA decomposed matrix is a highly dense matrix, so it is difficult to index individual dimension. LSA is unable to capture the multiple meanings of words. It offers lower accuracy than LDA.

In case of NMF, we observe same keywords are repeated in multiple topics. Keywords "go", "day" both are repeated in Topic 2, Topic 3, and Topic 4 (Table \ref{tab:topic_keywords}). In Table \ref{tab:topic_keywords} keyword "yoga" has been found both in Topic 1 and Topic 4. We also notice that keyword "eat" is in Topic 2 and Topic 3 (Table \ref{tab:topic_keywords}). If the same keywords being repeated in multiple topics, it is probably a sign that the `k' is large though we achieve the highest coherence score in NMF for k=4.

We use LDA model for our further analysis. Because LDA is good in identifying coherent topics where as NMF usually gives incoherent topics. However, in the average case NMF and LDA are similar but LDA is more consistent. 

\subsection{Topic Inference}
\label{subsec:3.4}
After doing topic modeling using three different method LSA, NMF, and LDA, we use LDA for further analysis i.e. to observe the dominant topic, 2$^{nd}$ dominant topic and percentage of contribution of the topics in each tweet of training data. To observe the model behavior on new tweets those are not included in training set, we follow the same procedure to observe the dominant topic, 2$^{nd}$ dominant topic and percentage of contribution of the topics in each tweet on testing data. Table \ref{tab:observ} shows some tweets and corresponding dominant topic, 2$^{nd}$ dominant topic and percentage of contribution of the topics in each tweet.

\subsection{Manual Annotation}
\label{subsec:3.5}
To calculate the accuracy of model in comparison with ground truth label, we selected top 500 tweets from train dataset (40k tweets). We extracted 500 new tweets (22 April, 2019) as a test dataset. We did manual annotation both for train and test data by choosing one topic among the 4 topics generated from LDA model (7$^{th}$, 8$^{th}$, 9$^{th}$, and 10$^{th}$ columns of Table \ref{tab:topic_keywords}) for each tweet based on the intent of the tweet. Consider the following two tweets:

Tweet 1: \textit{Learning some traditional yoga with my good friend.}

Tweet 2: \textit{Why You Should \#LiftWeights to Lose \#BellyFat \#Fitness \#core \#abs \#diet \#gym \#bodybuilding \#workout \#yoga}

The intention of Tweet 1 is yoga activity (i.e. learning yoga). Tweet 2 is more about weight lifting to reduce belly fat. This tweet is related to workout. When we do manual annotation, we assign Topic 2 in Tweet 1, and Topic 1 in Tweet 2. It's not wise to assign Topic 2 for both tweets based on the keyword "yoga". During annotation, we focus on functionality of tweets.

\begin{figure*}[t!]
\centering
    \begin{subfigure}[t]{0.5\textwidth}            
            \includegraphics[width=\textwidth]{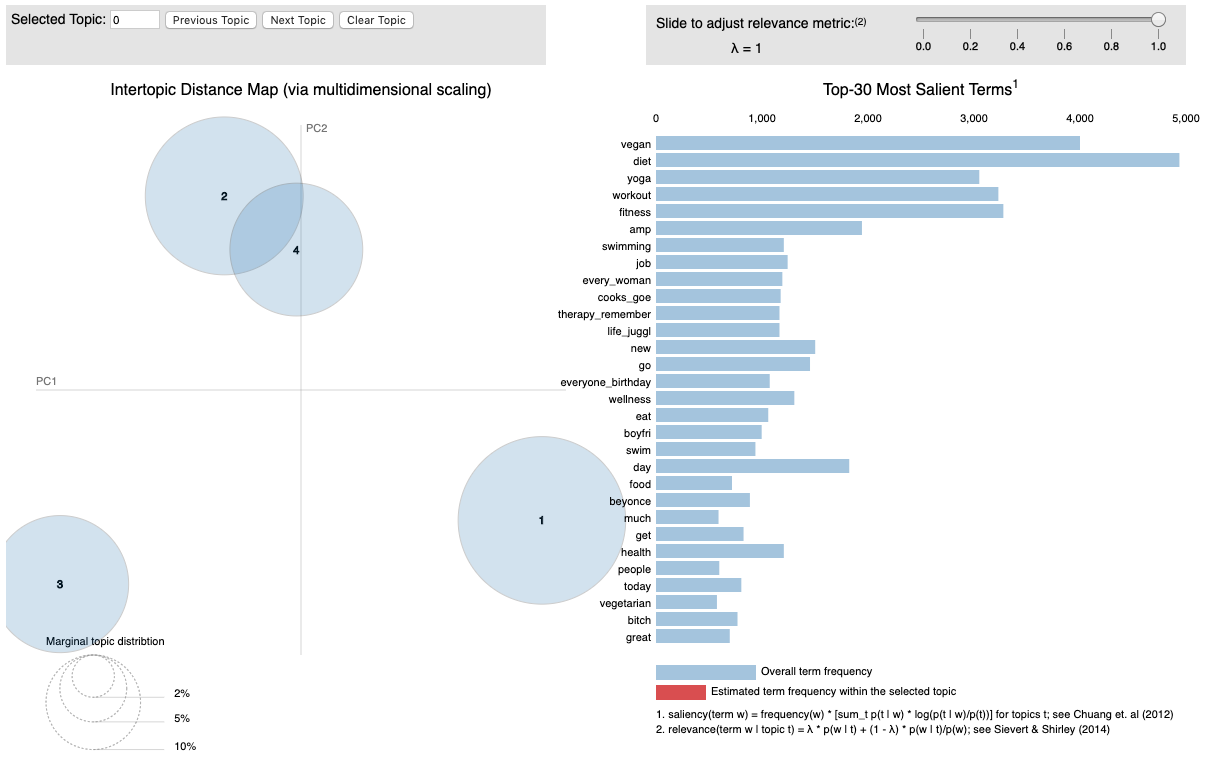}
            \caption{Bubbles in left hand side show overall topic distribution and sky blue bars in right hand side represent overall term frequencies. Best viewed in electronic format (zoomed in).}
            \label{fig:LDAVis1}
    \end{subfigure}
    ~
    \begin{subfigure}[t]{0.5\textwidth}
            \centering
            \includegraphics[width=\textwidth]{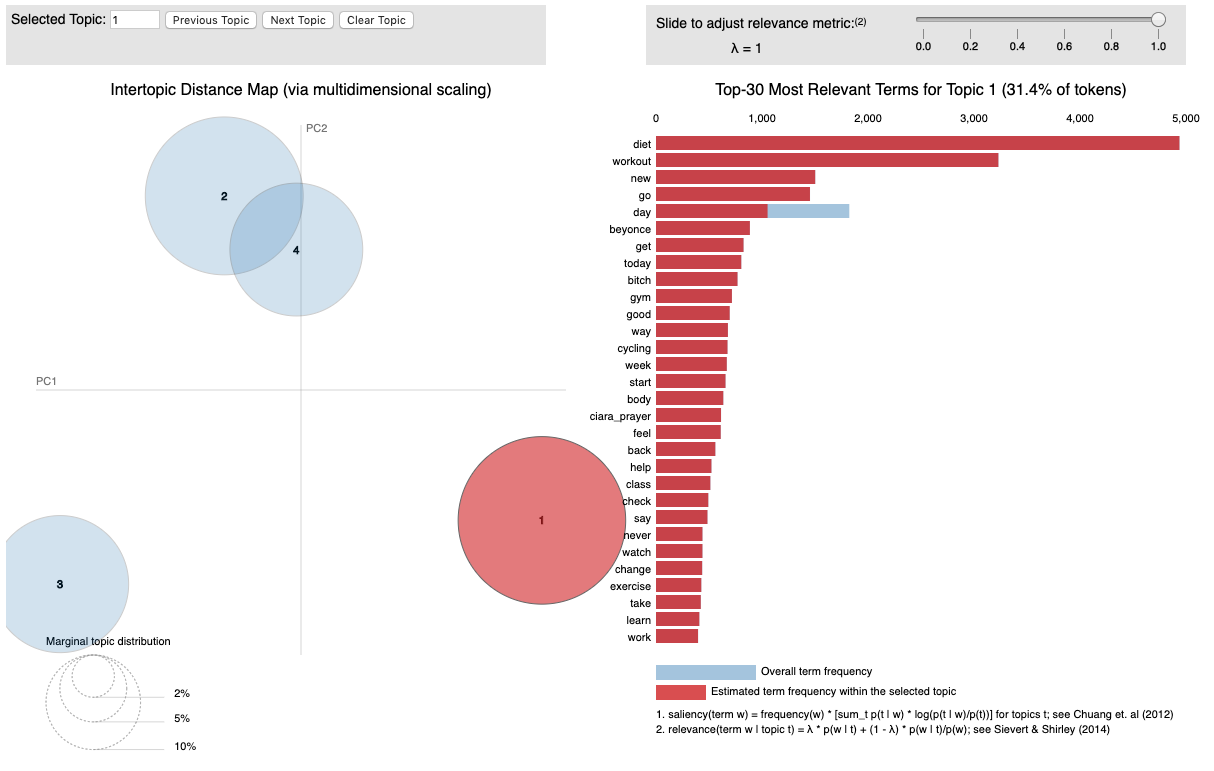}
            \caption{Red bubble in left hand side represents the selected Topic which is Topic 1. Red bars in right hand side show estimated term frequencies of top-30 salient keywords that form the  Topic 1. Best viewed in electronic format (zoomed in).}
            \label{fig:LDAVis2}
    \end{subfigure}
    \caption{Visualization using pyLDAVis. Best viewed in electronic format (zoomed in).}\label{fig:LDAVis}
\end{figure*}

\begin{figure}[htbp]
  \centering  
  \includegraphics[width= 0.5\textwidth]{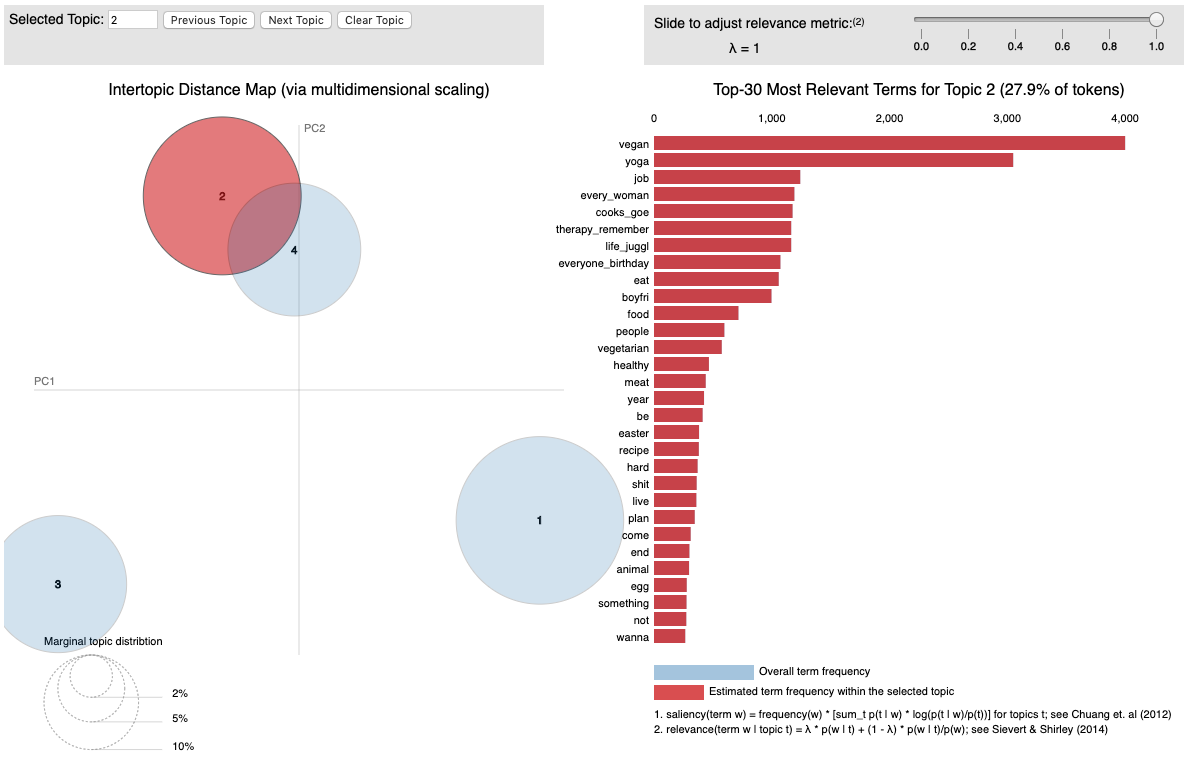}
    \caption{Visualization using pyLDAVis. Red bubble in left hand side represents selected Topic which is Topic 2. Red bars in right hand side show estimated term frequencies of top-30 salient keywords that form the Topic 2. Best viewed in electronic format (zoomed in)}
    \label{fig:LDAVis_T2}
\end{figure}

\begin{figure*}[t!]
\centering
    \begin{subfigure}[t]{0.5\textwidth}            
            \includegraphics[width=\textwidth]{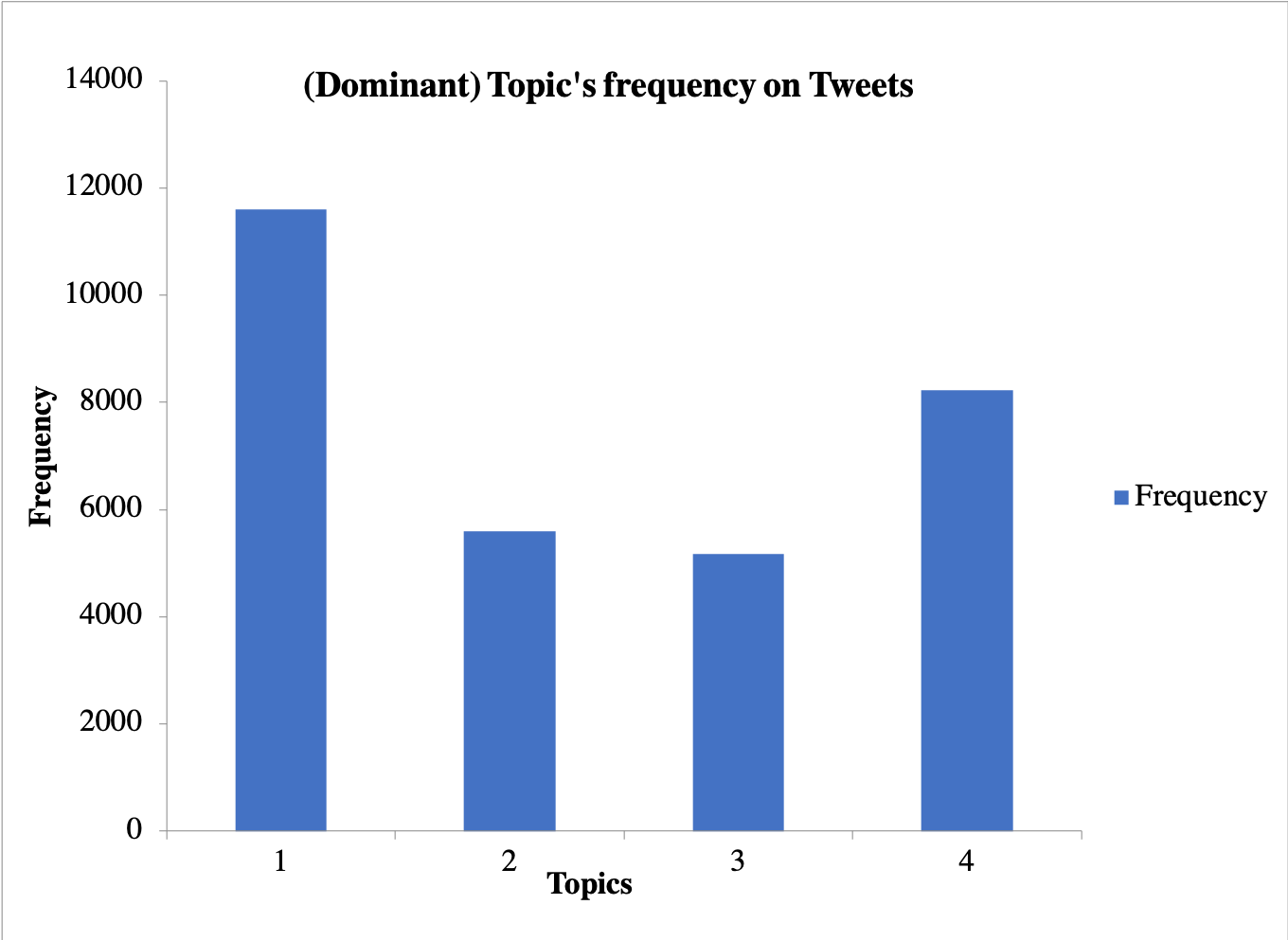}
            \caption{Dominant topic.}
            \label{fig:his1}
    \end{subfigure}
    ~
    \begin{subfigure}[t]{0.5\textwidth}
            \centering
            \includegraphics[width=\textwidth]{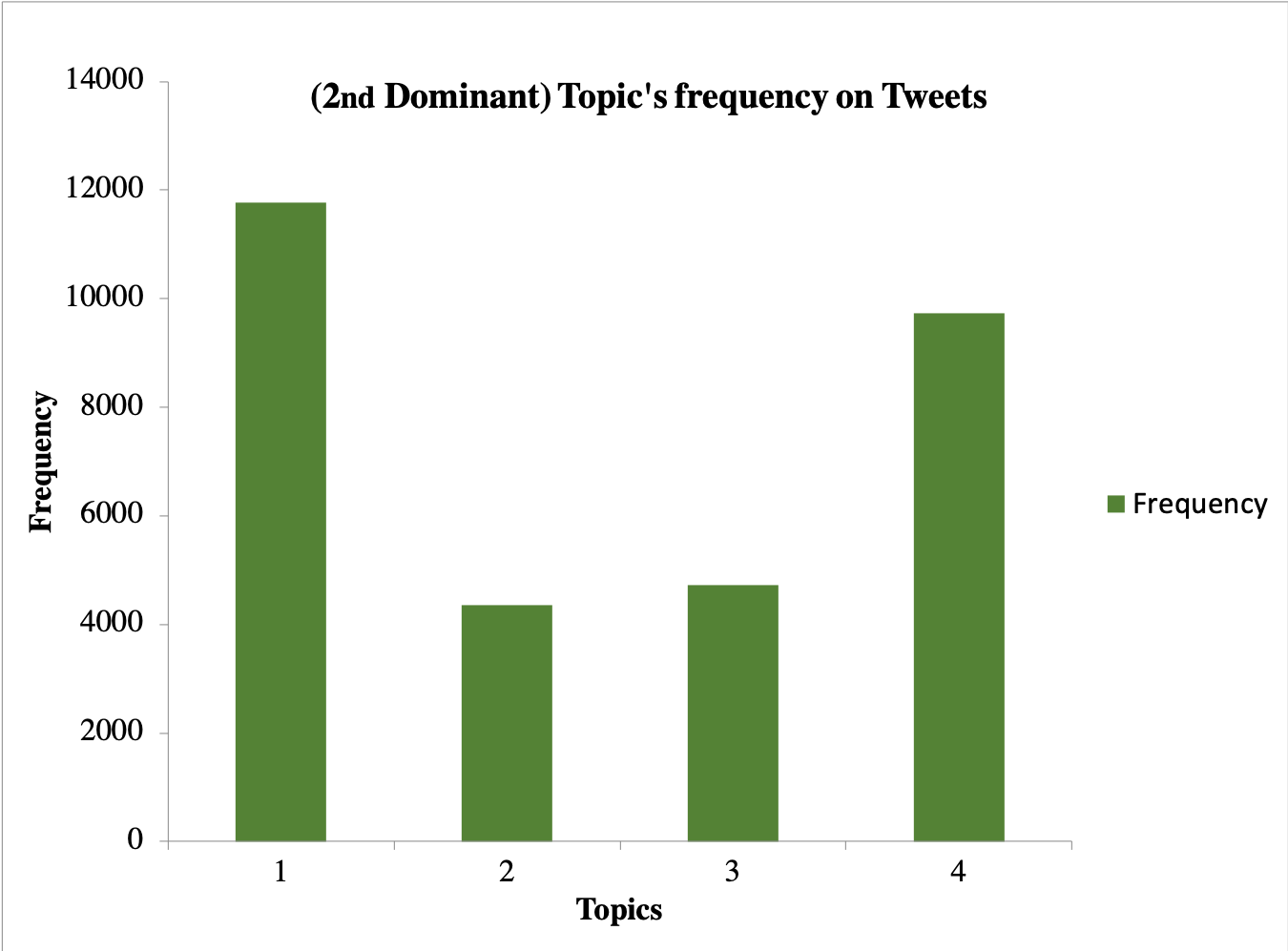}
            \caption{2$^{nd}$ dominant topic.}
            \label{fig:his2}
    \end{subfigure}
    \caption{Frequency of each topic's distribution on tweets.}\label{fig:histogram_topic}
\end{figure*}

\begin{figure}[htbp]
  \centering  
  \includegraphics[width= 0.5\textwidth]{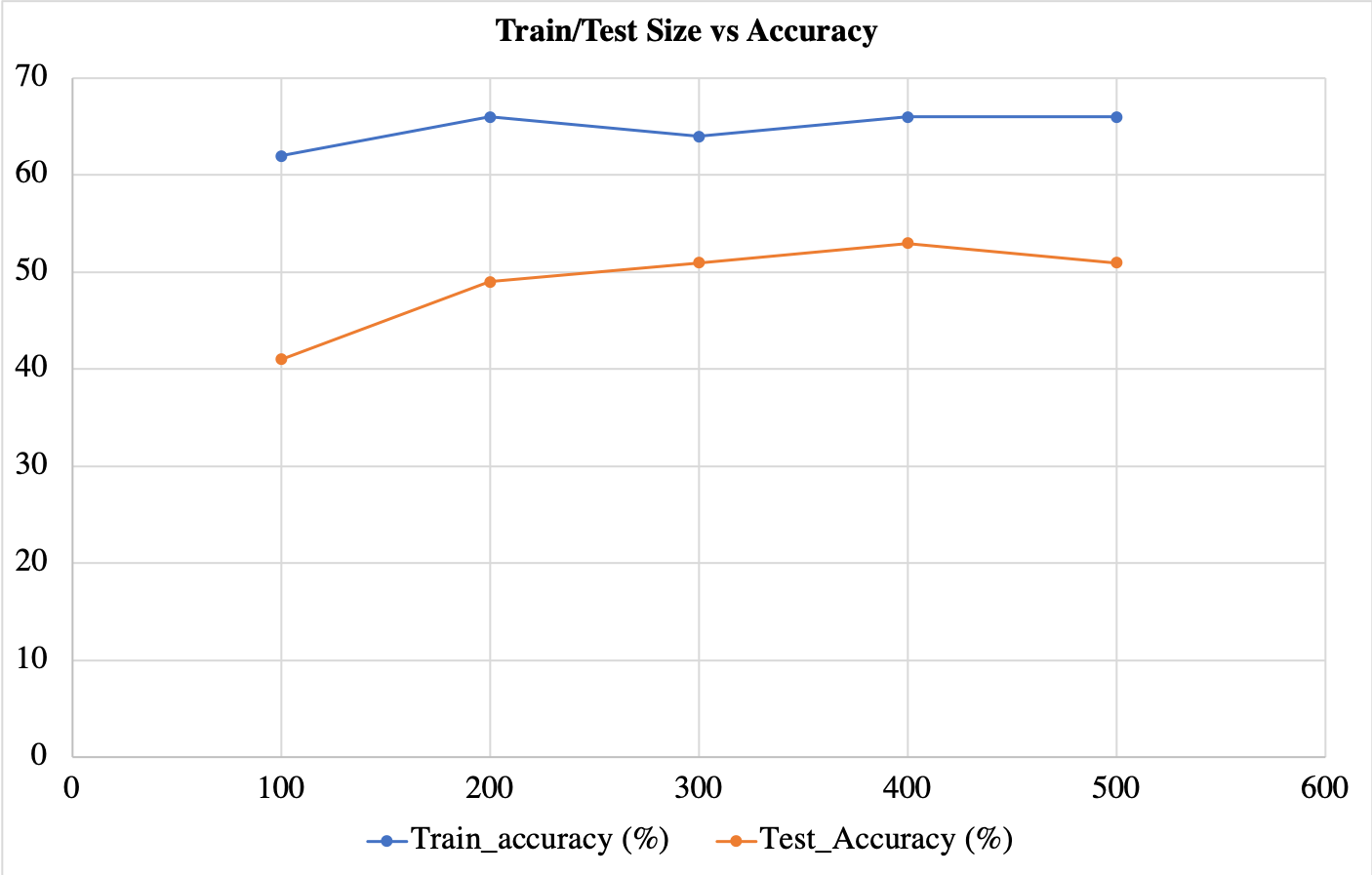}
    \caption{Percentage of Accuracy (y-axis) vs Size of Dataset (x-axis). Size of Dataset = 100, 200, 300, 400, and 500 tweets. Blue line shows the accuracy of Train data and Orange line represents Test accuracy. Best viewed in electronic format (zoomed in).}
    \label{fig:accuracy}
\end{figure}

\section{Results and Discussion}
\subsection{Visualization}
We use LDAvis \cite{sievert2014ldavis}, a web-based interactive visualization of topics estimated using LDA. Gensim's pyLDAVis is the most commonly used visualization tool to visualize the information contained in a topic model. In Fig. \ref{fig:LDAVis}, each bubble on the left-hand side plot represents a topic. The larger the bubble, the more prevalent is that topic. A good topic model has fairly big, non-overlapping bubbles scattered throughout the chart instead of being clustered in one quadrant.
A model with too many topics, is typically have many overlaps, small sized bubbles clustered in one region of the chart. In right hand side, the words represent the salient keywords.

If we move the cursor over one of the bubbles (Fig. \ref{fig:LDAVis2}), the words and bars on the right-hand side have been updated and top-30 salient keywords that form the selected topic and their estimated term frequencies are shown.

We observe interesting hidden correlation in data. Fig. \ref{fig:LDAVis_T2} has Topic 2 as selected topic. Topic 2 contains top-4 co-occurring keywords "vegan", "yoga", "job", "every\_woman" having the highest term frequency. We can infer different things from the topic that "women usually practice yoga more than men", "women teach yoga and take it as a job", "Yogi follow vegan diet". We would say there are noticeable correlation in data i.e. `Yoga-Veganism', `Women-Yoga'.

\begin{table*}[htbp]
  \centering
  \caption{The Dominant \& 2$^{nd}$ Dominant Topic of a Tweet and corresponding Topic Contribution on that specific Tweet.}
    \resizebox{\textwidth}{!}{\begin{tabular}{|c|p{19.665em}|c|c|c|c|}
    \hline
    \multicolumn{1}{|c|}{\textbf{Dataset}} & \multicolumn{1}{|c|}{\textbf{Tweets}} & \multicolumn{1}{|c}{\textbf{Dominant Topic}} & \multicolumn{1}{|c}{\textbf{Contribution (\%)}} & \multicolumn{1}{|c}{\textbf{2$^{nd}$ Dominant Topic}} & \multicolumn{1}{|c|}{\textbf{Contribution (\%)}} \\
    \hline
    Train & Revoking my vegetarian status till further notice. There's something I wanna do and I can't afford the supplements that come with being veggie. & 2     & 61    & 1     & 18 \\
    \hline
    Test  & I would like to take time to wish "ALL" a very happy \#EarthDay! \#yoga \#meditation & 2     & 33    & 4     & 32 \\
    \hline
    Test  & This morning I packed myself a salad. Went to yoga during lunch. And then ate my salad with water in hand. I'm feeling so healthy I don't know what to even do with myself. Like maybe I should eat a bag of chips or something. & 2     & 43    & 3     & 23 \\
    \hline
    Test  & My extra sweet halfcaf double vegan soy chai pumpkin latte was 2 degrees hotter than it should have been and the foam wasn't very foamy. And they spelled my name Jimothy, "Jim" on the cup. it's a living hell here. & 3     & 37    & 2     & 33 \\
    \hline
    \end{tabular}}
  \label{tab:observ}%
\end{table*}

\subsection{Topic Frequency Distribution}
Each tweet is composed of multiple topics. But, typically only one of the topics is dominant. We extract the dominant and 2$^{nd}$ dominant topic for each tweet and show the weight of the topic (percentage of contribution in each tweet) and the corresponding keywords. 

We plot the frequency of each topic's distribution on tweets in histogram. Fig. \ref{fig:his1} shows the dominant topics' frequency and Fig. \ref{fig:his2} shows the 2$^{nd}$ dominant topics' frequency on tweets. From Fig. \ref{fig:histogram_topic} we observe that Topic 1 became either the dominant topic or the 2$^{nd}$ dominant topic for most of the tweets. 7$^{th}$ column of Table \ref{tab:topic_keywords} shows the corresponding top-10 keywords of Topic 1. 

\subsection{Comparison with Ground Truth}
To compare with ground truth, we gradually increased the size of dataset 100, 200, 300, 400, and 500 tweets from train data and test data (new tweets) and did manual annotation both for train/test data based on functionality of tweets (described in \hyperref[subsec:3.5]{Subsection 3.5}). 

For accuracy calculation, we consider the dominant topic only. We achieved 66\% train accuracy and 51\% test accuracy when the size of dataset is 500 (Fig. \ref{fig:accuracy}). We did baseline implementation with random inference by running multiple times with different seeds and took the average accuracy. For dataset 500, the accuracy converged towards 25\% which is reasonable as we have 4 topics.

\subsection{Observation and Future Work}
In Table \ref{tab:observ}, we show some observations. For the tweets in 1$^{st}$ and 2$^{nd}$ row (Table \ref{tab:observ}), we observed understandable topic. We also noticed misleading topic and unrelated topic for few tweets (3$^{rd}$ and 4$^{th}$ row of Table \ref{tab:observ}).

In the 1$^{st}$ row of Table \ref{tab:observ}, we show a tweet from train data and we got Topic 2 as a dominant topic which has 61\% of contribution in this tweet. Topic 1 is 2$^{nd}$ dominant topic and 18\% contribution here.  

2$^{nd}$ row of Table \ref{tab:observ} shows a tweet from test set. We found Topic 2 as a dominant topic with 33\% of contribution and  Topic 4 as 2$^{nd}$ dominant topic with 32\% contribution in this tweet.

In the 3$^{rd}$ (Table \ref{tab:observ}), we have a tweet from test data and we got Topic 2 as a dominant topic which has 43\% of contribution in this tweet. Topic 3 is 2$^{nd}$ dominant with 23\% contribution which is misleading topic. The model misinterprets the words `water in hand' and infers topic which has keywords "swimming, swim, pool". But the model should infer more reasonable topic (Topic 1 which has keywords "diet, workout") here.

We got Topic 2 as dominant topic for the tweet in 4$^{th}$ row (Table \ref{tab:observ}) which is unrelated topic for this tweet and most relevant topic of this tweet (Topic 2) as 2$^{nd}$ dominant topic. We think during accuracy comparison with ground truth 2$^{nd}$ dominant topic might be considered.

In future, we will extract more tweets and train the model and observe the model behavior on test data. As we found misleading and unrelated topic in test cases, it is important to understand the reasons behind the predictions. We will incorporate Local Interpretable model-agnostic Explanation (LIME) \cite{ribeiro2016should} method for the explanation of model predictions. We will also do predictive causality analysis on tweets.

\section{Conclusions}
It is challenging to analyze social media data for different application purpose. In this work, we explored Twitter health-related data, inferred topic using topic modeling (i.e. LSA, NMF, LDA), observed model behavior on new tweets, compared train/test accuracy with ground truth, employed different visualizations after information integration and discovered interesting correlation (Yoga-Veganism) in data. In future, we will incorporate Local Interpretable model-agnostic Explanation (LIME) method to understand model interpretability.

\bibliographystyle{ACM-Reference-Format}
\bibliography{acmart}

\end{document}